# Ray-Based and Graph-Based Methods for Fiber Bundle Boundary Estimation

M. H. A. Bauer, J. Egger, D. Kuhnt, S. Barbieri, J. Klein, H.-K. Hahn, B. Freisleben, Ch. Nimsky

*Abstract*—Diffusion Tensor Imaging (DTI) provides the possibility of estimating the location and course of eloquent structures in the human brain. Knowledge about this is of high importance for preoperative planning of neurosurgical interventions and for intraoperative guidance by neuronavigation in order to minimize postoperative neurological deficits. Therefore, the segmentation of these structures as closed, three-dimensional object is necessary. In this contribution, two methods for fiber bundle segmentation between two defined regions are compared using software phantoms (abstract model and anatomical phantom modeling the right corticospinal tract). One method uses evaluation points from sampled rays as candidates for boundary points, the other method sets up a directed and weighted (depending on a scalar measure) graph and performs a min-cut for optimal segmentation results. Comparison is done by using the Dice Similarity Coefficient (DSC), a measure for spatial overlap of different segmentation results.

## I. INTRODUCTION

*Diffusion Tensor Imaging (DTI)* is a non-invasive imaging modality that facilitates the estimation of location and course of white matter tracts in the human brain in-vivo. This information about eloquent structures is of main importance for neurosurgical interventions. Major white matter tracts like the corticospinal tract have to be protected during intervention to avoid preoperative neurological deficits. Therefore, the eloquent structures are reconstructed from DTI data for preoperative planning and intraoperative visualization in the operating room microscope. For this purpose a corresponding 3D object has to be created to visualize the bounding curves, that show the extent of the fiber bundle, which is indispensable for the intervention [1], [2].

DTI allows the estimation of white matter tracts based on the measurement of water diffusion. Based on diffusion-weighted pulse sequences (at least six diffusion-weighted images with different gradient directions besides one unweighted image) that are sensitive to the random motion of the water molecules, 2nd order tensors can be calculated for each voxel, described by the symmetric diffusion tensor matrix D:

$$D = \begin{pmatrix} D_{xx} & D_{xy} & D_{xz} \\ D_{yx} & D_{yy} & D_{yz} \\ D_{zx} & D_{zy} & D_{zz} \end{pmatrix} \qquad (1)$$

Three eigenvectors $(\vec{e}_1, \vec{e}_2, \vec{e}_3)$ and the corresponding eigenvalues $(\lambda_1, \lambda_2, \lambda_3)$ can be calculated by diagonalizing D, describing the three main diffusion directions and its magnitude [3], [4], [5]. Depending on this characteristics, different scalar anisotropy measures can be defined like the fractional anisotropy (FA) [6]. The FA value describes the fraction of the "magnitude" of $D$ that is ascribed to the anisotropic diffusion [3] and is mentioned by [7].

$$FA = \sqrt{\frac{(\lambda_1 - \lambda_2)^2 + (\lambda_2 - \lambda_3)^2 + (\lambda_1 - \lambda_3)^2}{2 \cdot (\lambda_1^2 + \lambda_2^2 + \lambda_3^2)}} \qquad (2)$$

Depending on the tensor and the derived measures two algorithms for fiber bundle segmentation have been developed and will be evaluated against each other.

There exist some approaches to reconstruct white matter tracts and to visualize a closed segmented 3D object. Based on fiber tracking, which only delivers a set of streamlines with no border information, hulls can be generated to overcome this limitation. This can be done in several ways like described in [2], [8] by generating a surface that wraps the computed streamlines with the help of bounding curves along the fiber tracking result. Another method was presented by [4], describing a directed volume growing approach.

This contribution is organized as follows. Section II describes the two developed segmentation algorithms. In Section III both approaches are compared against each other. Section IV concludes the paper and outlines areas for future work.

## II. METHODS

### A. Preprocessing

Both methods depend on the same preprocessing step for segmentation of a fiber bundle. With the help of a user defined and manually placed seed region (as 2D contour) a deterministic fiber tracking algorithm (sampled 2D contour delivers set of seed points) is applied to reconstruct an initial set of fiber tracts. With the help of two include regions the tracking results are restricted to the structure of interest. Based on the tracked fibers the centerline of the bundle is calculated by averaging the sampled fibers, as described by [9]. The calculated centerline is now sampled at *n* points (see Figure 1).

M.H.A.B., J.E., D.K., Ch.N. are with the Department of Neurosurgery, University of Marburg, Marburg, Germany e-mail: bauermi@med.uni-marburg.de
M.H.A.B., J.E., B.F. are with the Department of Mathematics and Computer Science, University of Marburg, Marburg, Germany
S.B., J.K., H.-K.H. are with Fraunhofer MEVIS – Institute for Medical Image Computing, Bremen, Germany

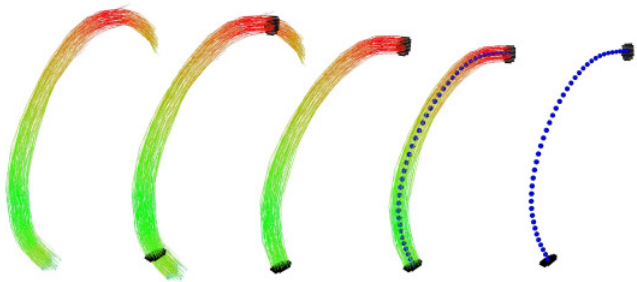

Fig. 1. Preprocessing part one (from left to right): (1) Fiber tracking with given seed region, (2) manually placing of include regions, (3) cropping of tracking result, (4) centerline calculation, (5) centerline sampling between include regions.

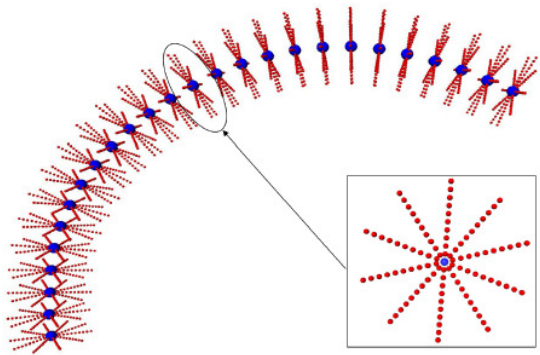

Fig. 2. Preprocessing part two: plane calculation along centerline and sending out rays for evaluation point determination.

For each point a plane is generated upright to the centerline's direction given by the difference of two consecutive centerline points. Within each plane, k equidistant distributed vectors are send out and each of them is sampled at $m$ equally spaced points (with distance $d$). This results in a set of $n * k * m$ evaluation points for segmentation (see Figure 2).

### B. Ray-based Segmentation

For the ray-based segmentation some information derived from the diffusion tensors is determined for each evaluation point:

- FA value
- angle $\alpha_c$ between main diffusion direction of the corresponding (in plane) centerline point and the main diffusion direction of the evaluation point
- angle $\alpha_n$ between the main diffusion direction of the evaluation point and the previous evaluation point along the ray

Now every ray is analyzed using the mentioned parameters FA value, $\alpha_c$ and $\alpha_n$ with the help of threshold criteria, concerning the actual evaluation point and $r$ previous (if possible) evaluation points along the ray, where $r$ is given by $r \geq voxeldiagonal/d$. This guarantees that at least two voxels are considered for the decision to reduce the influence of noise.

Afterwards, different postprocessing steps can be used for smoothing the calculated point cloud. For this purpose an in-plane correction scheme can be applied, that prevents the single 2D contours from extreme outliers. For intra-plane correction the found boundary points of the rays with the same index of consecutive planes are considered and corrected to have similar distances to the centerline. For further details see [10], [11].

### C. Graph-based Segmentation

The graph-based segmentation method uses up to now only the FA value of the evaluation points for the decision of belonging to the fiber bundle or not. For this purpose a directed and weighted graph $G = (V,E)$ is set up, whereas the construction is based on the methods introduced in [12], [13], [14]. Besides all evaluation points as nodes $v \in V$ of the graph, two additional nodes $s$ and $t$ are inserted as source and sink. The edges $e \in E$ are connecting respectively two nodes. The weights are set to maximum weight ($\infty$) for edges connecting the points of one ray (see Figure 3 top), the points of the same rays of neighbored planes (see Figure 3 bottom) and the points of neighbored rays of the same plane (see Figure 3 middle) using additionally defined parameters for smoothness. Edges connecting the points to sink and source are weighted depending on the FA value of the point and the average FA value of the fiber bundle. For further details see [15].

After graph construction a minimal cost closed set is computed via a polynomial time s-t-cut [16], delivering an optimal segmentation (depending on the weightening function) of the fiber bundle as set of boundary points.

### D. Postprocessing

In order to use the found boundary, given as set of boundary points, for preoperative planning and intraoperative visualization a 3D object has to be created. Due to the point order, given by the planes and rays, two neighbored 2D contour point sets are triangulated like shown in Figure 4. Finally, the first and last 2D contour are triangulated by using the centerline point of the corresponding plane respectively to form a closed 3D object.

### E. Data

For evaluation and comparison of the two approaches software phantoms are used, to have ground truth data to compare against. As first software phantom a portion of a torus with a voxel size of $1 \times 1 \times 1 mm^3$ and a diameter of $10mm$ of the area cross-section [17] is used. As first step towards more anatomical data, another software phantom (see Figure 5) with similar resolution is used [18]. Besides a modeled right corticospinal tract also several tissues like white matter, gray matter and cerebrospinal fluid are modeled.

### III. RESULTS

Both methods were implemented in C++ within the medical platform MeVisLab [19] and were executed on an Intel Core 2 Quad CPU, 3GHz, 6 GB RAM, Windows XP Professional 2003, SP 2.

With the knowledge of dimension and location of the modeled fiber tracts of the software phantoms, the cutout of the model between the two manually places regions and the calculated contour given by the two algorithms are compared. For this purpose the given reference fiber bundle and the segmented fiber bundle are used. For comparison, the *Dice Similarity Coefficient (DSC)* was used [20], [21]. The DSC is commonly used in medical imaging studies in order to quantify the degree of overlap between two segmented objects A and B and is given by:

$$DSC = \frac{2 \cdot |A \cap B|}{|A| + |B|} \qquad (3)$$

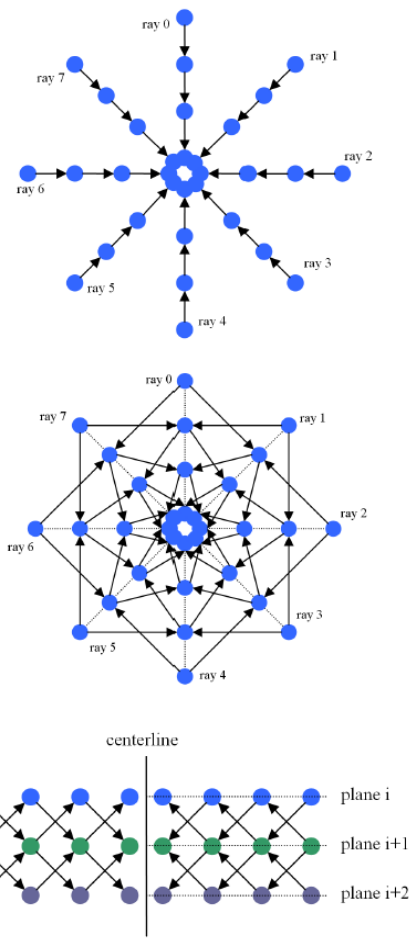

Fig. 3. Principle of ∞-weighted edge construction: along single rays (top), between neighboring planes with smoothness parameter 1 (middle) and between neighboring rays of the same plane with smoothness parameter 1 (bottom).

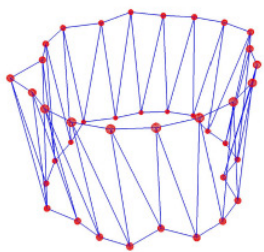

Fig. 4. Triangulation scheme for two neighbored point sets of 2D contours.

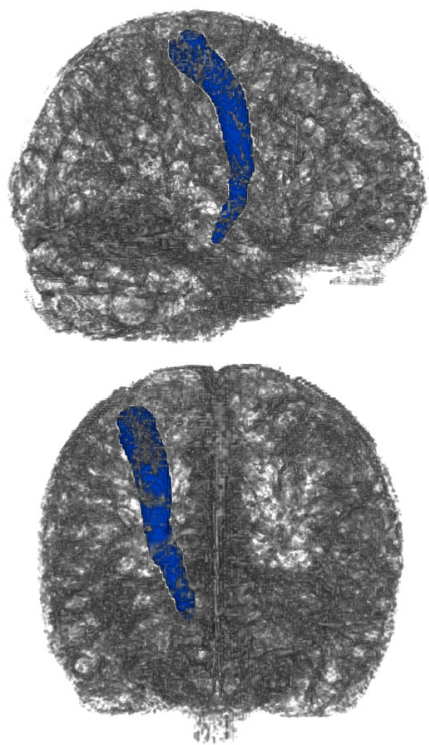

Fig. 5. Brainweb-based software phantom [18] with modeled right corti-cospinal tract (blue): view from the right side (top) and view from the front (bottom)

TABLE I
EVALUATION RESULTS OF BOTH SEGMENTATION APPROACHES FOR THE
TORUS SHAPED SOFTWARE PHANTOM (TOP) AND THE ANATOMICAL
SOFTWARE PHANTOM WITH MODELED CORTICOSPINAL TRACT (BOTTOM).

| phantom 1 | ray-based approach | graph-based approach |
|---|---|---|
| min DSC(%) | 74.721 | 68.330 |
| max DSC(%) | 91.532 | 78.138 |
| average DSC(%) | 88.462 | 74.171 |
| standard deviation | 4.438 | 3.999 |

| phantom 2 | ray-based approach | graph-based approach |
|---|---|---|
| min DSC(%) | 75.321 | 63.142 |
| max DSC(%) | 88.502 | 80.747 |
| average DSC(%) | 81.538 | 73.731 |
| standard deviation | 4.918 | 5.119 |

Both approaches were applied to the given software-phantoms, the portion of torus and the anatomical phantom with modeled corticospinal tract, with different parameter configurations like smoothness control within the graph-based approach or application of correction schemes within the ray-based approach. The evaluation of the created segmentations induced a mean DSC for the ray-based approach of 85.00% while the graph-based approach yielded a DSC of 73.95% like given in Table I. An example of boundary estimation with the help of the graph-based approach is shown in Figure 6, where a part of the right corticospinal tract is modeled.

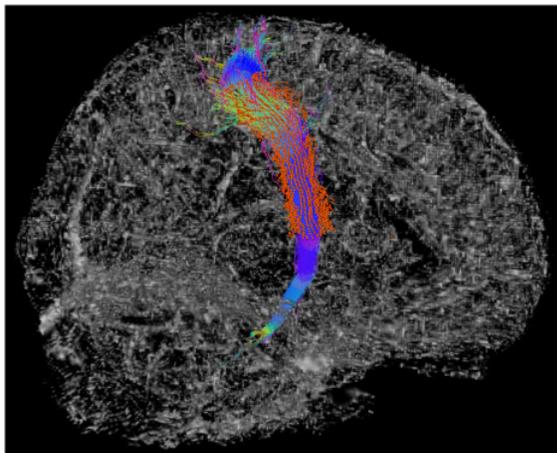

Fig. 6. Segmentation result (boundary point cloud) for the second software phantom describing the modeled right corticospinal tract.

## IV. Conclusion

In this paper, two approaches for determination of the fiber bundle boundary were introduced and compared to each other. Both approaches use a centerline derived from initial tracking between two manually placed regions for identifying the structure of interest. After sampling the centerline and the calculation of planes up right to the centerlines direction, rays are sent out and sampled within each plane, creating a set of evaluation points. The ray-based approach searches for boundary points along each ray. The graph-based method sets up a directed and weighted graph and calculates a min-cut, delivering an optimal segmentation, based on the given cost function. Both approaches result in a set of boundary points that were triangulated for 3D object generation.

Both approaches were evaluated against each other using software phantoms by calculating the Dice Similarity Coefficient showing a higher DSC for the ray-based approach (85.00%) in contrast to the graph-based method (73.95%) depending on only one scalar measure.

There are several areas of future work. In the case of the ray-based approach additional tensor information (other scalar measures directed diffusion [7]) could be considered for boundary estimation along the rays or the area of interest for each decision could be enhanced to neighboring rays and planes instead of single rays. In case of the graph-based method an extended cost function for graph-weighting considering direction information and other scalar measures would be a possible extension.

## References

[1] D. Merhof, M. Meister, E. Bingl, Ch. Nimsky, and G. Greiner, "Isosurface-Based Generation of Hulls Encompassing Neuronal Pathways," *Stereotact Funct Neurosurg*, vol. 87, pp. 50–60, 2009.

[2] Ch. Nimsky, O. Ganslandt, F. Enders, D. Merhof, T. Hammen, and M. Buchfelder, "Visualization strategies for major white matter tracts identified by diffusion tensor imaging for intraoperative use," in *Computer Assisted Radiology and Surgery*, vol. 1281, 2005, pp. 793 – 797.

[3] D. L. Bihan, J.-F. Mangin, C. Poupon, C. A. Clark, S. Pappata, N. Molko, and H. Chabriat, "Diffusion Tensor Imaging: Concepts and Applications," *Journal of Magnetic Resonance Imaging*, vol. 13, pp. 534–546, 2001.

[4] D. Merhof, P. Hastreiter, Ch. Nimsky, R. Fahlbusch, and G. Greiner, "Directional Volume Growing for the Extraction of White Matter Tracts from Diffusion Tensor Data," in *SPIE - Medical Imaging 2005: Visualization, Image-Guided Procedures, and Display.*, vol. 5744, 2005, pp. 165–172.

[5] Y. Ou and C. Wyatt, "Visualization of Diffusion Tensor Imaging data and Image Correction of Distortion Induced by Patient Motion and Magnetic Field Eddy Current," in *Virginia Tech - Wake Forest University School of Biomedical Engineering and Sciences 4th Student Research Symposium*, 2005.

[6] P. J. Basser and C. Pierpaoli, "Microstructural and Physiological Features of Tissues Elucidated by Quantitative-Diffusion-Tensor MRI," *Journal of Magnetic Resonance, Series B*, vol. 111, no. 3, pp. 209–219, June 1996.

[7] T. H. J. M. Peeters, P. R. Rodrigues, A. Vilanova, and B. M. ter Haar Romeny, *Visualization and Processing of Tensor Fields*. Springer Berlin Heidelberg, 2009, ch. Analysis of Distance/Similarity Measures for Diffusion Tensor Imaging, pp. 113–136.

[8] Z. Ding, J. C. Gore, and A. W. Anderson, "Case study: reconstruction, visualization and quantification of neuronal fiber pathways," in *VIS '01: Proceedings of the conference on Visualization '01*. IEEE Computer Society, 2001, pp. 453–456.

[9] J. Klein, S. Hermann, O. Konrad, H. K. Hahn, and H.-O. Peitgen, "Automatic Quantification of DTI Parameters along Fiber Bundles," in *Proceeding of Image Processing for Medicine (BVM 2007)*, 2007, pp. 272–2.

[10] M. H. A. Bauer, S. Barbieri, J. Klein, J. Egger, D. Kuhnt, H. K. Hahn, B. Freisleben, and Ch. Nimsky, "A Ray-based Approach for Boundary Estimation of Fiber Bundles Derived from Diffusion Tensor Imaging," in *Proceedings of Computer Assisted Radiology and Surgery (CARS 2010)*, Geneva, Switzerland, June 2010.

[11] M. H. A. Bauer, S. Barbieri, J. Klein, J. Egger, D. Kuhnt, H.-K. Hahn, B. Freisleben, and Ch. Nimsky, "Boundary Estimation of Fiber Bundles derived from Diffusion Tensor Images," *International Journal of Computer Assisted Radiology and Surgery (IJCARS)*, 2010.

[12] K. Li, X. Wu, Z. Chen, and M. Sonka, "Optimal Surface Segmentation in Volumetric Images-A Graph-Theoretic Approach," *IEEE Transactions on Pattern Analysis and Machine Intelligence*, vol. 28, no. 1, pp. 119–134, 2006.

[13] J. Egger, T. O'Donnell, C. Hopfgartner, and B. Freisleben, "Graph-Based Tracking Method for Aortic Thrombus Segmentation," in *Proceedings of 4th European Congress for Medical and Biomedical Engineering, Engineering for Health*, Antwerp, Belgium, November 2008.

[14] J. Egger, B. Freisleben, R. Setser, R. Renapuraar, C. Biermann, and T. O'Donnell, "Aorta Segmentation for Stent Simulation," in *12th International Conference on MICCAI, Cardiovascular Interventional Imaging and Biophysical Modelling Workshop*, London, United Kingdom, 2009.

[15] M. H. A. Bauer, J. Egger, T. O'Donnell, S. Barbieri, J. Klein, B. Freisleben, H. K. Hahn, and Ch. Nimsky, "A Fast and Robust Graph-based Approach for Boundary Estimation of Fiber Bundles Relying on Fractional Anisotropy Maps," in *Proceedings of the 20th International Conference on Pattern Recognition (ICPR 2010)*, Istanbul, Turkey, August 2010.

[16] Y. Boykov and V. Kolmogorov, "An Experimental Comparison of Min-Cut/Max-Flow Algorithms for Energy Minimization in Vision," *IEEE Transactions on Pattern Analysis and Machine Intelligence*, vol. 26, pp. 359–374, 2001.

[17] S. Barbieri, J. Klein, Ch. Nimsky, and H.-K. Hahn, "Towards Image-Dependent Safety Hulls for Fiber Tracking," in *Proceedings of the Joint Annual Meeting ISMRM-ESMRMB 2010*, Stockholm, Sweden, 2010, p. 1672.

[18] S. Barbieri, J. Klein, Ch. Nimsky, and H. K. Hahn, "Assessing Fiber Tracking Accuracy via Diffusion Tensor Software Models," in *Medical Imaging 2010: Image Processing*, B. M. Dawant and D. R. Haynor, Eds., vol. 7623, no. 1. San Diego, California, USA: SPIE, 2010, p. 762326.

[19] http://www.mevislab.de.

[20] K. H. Zou, S. K. Warfield, A. Bharatha, C. M. C. Tempany, M. R. Kaus, S. J. Haker, W. M. Wells, F. A. Jolesz, and R. Kikinis, "Statistical Validation of Image Segmentation Quality Based on a Spatial Overlap Index," *Academic Radiology*, vol. 11, no. 2, pp. 178–189, 2004.

[21] M. P. Sampat, Z. Wang, M. K. Markey, G. J. Whitman, T. W. Stephens, and A. C. Bovik, "Measuring intra- and inter-oberserver Agreement in Identifying and Localizing Structures in Medical Images," *IEEE Inter. Conf. Image Processing*, 2006.